\documentclass[conference]{IEEEtran}
\IEEEoverridecommandlockouts
\usepackage{cite}
\usepackage{amsmath,amssymb,amsfonts}
\usepackage{algorithmic}
\usepackage{graphicx}
\usepackage{textcomp}
\usepackage{xcolor}
\usepackage{algorithm}
\usepackage{array}
\usepackage{stfloats}
\usepackage{url}
\usepackage{hyperref}
\usepackage{verbatim}

\usepackage{cite}
\usepackage{booktabs}
\usepackage{adjustbox}
\usepackage{multirow}
\usepackage{subcaption}
\usepackage{makecell}
\usepackage{soul}
\usepackage{cleveref}
\usepackage{xcolor}
\usepackage{colortbl}

\def\BibTeX{{\rm B\kern-.05em{\sc i\kern-.025em b}\kern-.08em
    T\kern-.1667em\lower.7ex\hbox{E}\kern-.125emX}}

\begin{document}

\title{Are Multimodal LLMs Ready for Surveillance? \\ A Reality Check on Zero-Shot Anomaly Detection in the Wild
}

\author{
    \IEEEauthorblockN{Shanle Yao\textsuperscript{*}, Armin Danesh Pazho\textsuperscript{*}, Narges Rashvand, and Hamed Tabkhi}
    \IEEEauthorblockA{\textit{Electrical and Computer Engineering Department} \\
    \textit{University of North Carolina at Charlotte}\\
    Charlotte, USA \\
    \{syao, adaneshp, nrashvan, htabkhiv\}@charlotte.edu}
    \thanks{*Both authors contributed equally to this research.}
}

\maketitle

\begin{abstract}
Multimodal large language models (MLLMs) have demonstrated impressive general competence in video understanding, yet their reliability for real-world Video Anomaly Detection (VAD) remains largely unexplored. Unlike conventional pipelines relying on reconstruction or pose-based cues, MLLMs enable a paradigm shift: treating anomaly detection as a language-guided reasoning task. In this work, we systematically evaluate state-of-the-art MLLMs on the ShanghaiTech and CHAD benchmarks by reformulating VAD as a binary classification task under weak temporal supervision. We investigate how prompt specificity and temporal window lengths (1s–3s) influence performance, focusing on the precision–recall trade-off. Our findings reveal a pronounced conservative bias in zero-shot settings; while models exhibit high confidence, they disproportionately favor the 'normal' class, resulting in high precision but a recall collapse that limits practical utility. We demonstrate that class-specific instructions can significantly shift this decision boundary—improving the peak F1-score on ShanghaiTech from 0.09 to 0.64—yet recall remains a critical bottleneck. These results highlight a significant performance gap for MLLMs in noisy environments and provide a foundation for future work in recall-oriented prompting and model calibration for open-world surveillance, which demands complex video understanding and reasoning. 
\end{abstract}


\section{Introduction}
\label{sec:intro}

The landscape of computer vision is being rapidly reshaped by Multimodal Large Language Models (MLLMs), which have moved beyond static recognition toward joint visual–language reasoning over complex video streams. Recent efforts reflect this shift. LVBench \cite{wang2025lvbench} pushes models toward extreme long-video understanding, requiring temporal coherence over minutes to hours. MMBench-Video \cite{fang2024mmbench} further broadens evaluation by testing free-form questions and temporal reasoning that better resemble practical interactions. Together, these benchmarks suggest that systems such as GPT-4V and Gemini are evolving from visual classifiers into general-purpose reasoning engines capable of interpreting dynamic visual narratives.

Yet the question of whether these capabilities translate into operational reliability remains open, especially for Video Anomaly Detection (VAD), a domain where errors carry real consequences. VAD is central to community safety and intelligent surveillance, aiming to detect irregular events such as accidents, intrusions, or public disturbances \cite{noghre2025survey}. As public spaces increasingly rely on automated monitoring, detection systems must not only “see motion,” but also reason about intent, context, and risk. In principle, MLLMs are well-positioned for this: their semantic priors and language interface could enable more interpretable detection decisions, richer explanations, and scalable annotation, capabilities that classical reconstruction, or pose-based pipelines do not natively provide\cite{pazho2023ancilia, yao2025lab}.

\begin{figure}[]
    \centering
    \includegraphics[clip,trim={0 125 100 10},width=1\columnwidth]{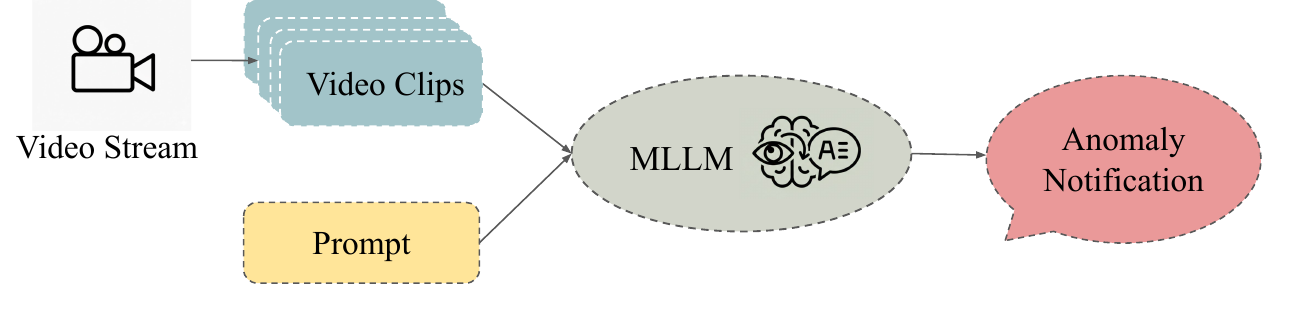}
    \caption{Conceputal Overview}
    \label{fig:conceptual}
\end{figure}

However, moving from controlled benchmarks to real-world VAD introduces challenges that are not well captured by standard evaluation protocols. Real surveillance footage is noisy and semantically ambiguous, and the meaning of “anomaly” is inherently contextual: running may be normal in a park but suspicious in a restricted area. Moreover, many VAD studies report ranking-oriented metrics such as AUC-ROC or AUC-PR, which are useful for offline comparison but do not directly yield the actionable decision boundaries required by live systems.\cite{yao2024evaluating} This creates a practical “decision gap”: operators must choose thresholds without a principled criterion for when an alert should be triggered. Recent work has begun to narrow this gap using human-in-the-loop strategies \cite{yao2025alfred}, for example by updating thresholds through active learning. Still, an unresolved question remains: can general-purpose MLLMs serve as dependable anomaly decision-makers under open-world uncertainty, without extensive task-specific training?

A key reason this question remains under-explored is the lack of evaluation settings that truly match open-ended, real surveillance footage. Much of what we know about MLLM video competence comes from curated action-oriented datasets—often trimmed, staged, or captured with relatively clean viewpoints and consistent semantics.\cite{liu2019ntu} In contrast, surveillance cameras typically capture longer temporal context, lower-resolution subjects, occlusions, crowding, and frequent background distractions, while anomalies may be rare, ambiguous, and only identifiable through subtle context over time\cite{yao2026offline, pazho2025towards}. At the other end of the spectrum, movies or online videos often contain strong narrative structure and cinematographic cues, shot composition, edits, and implied intent, that can implicitly guide interpretation in ways that do not exist in fixed-camera monitoring. Bridging this gap requires attention: without surveillance-realistic protocols, it is difficult to determine whether apparent “video understanding” reflects robust reasoning or reliance on dataset-specific regularities.

To study this problem, we introduce a framework that integrates state-of-the-art MLLMs directly into the VAD pipeline as prompt-conditioned decision modules. Rather than treating anomaly detection as reconstruction failure or pose deviation, we reformulate VAD as a prompt-guided binary classification task, where the model must decide whether a short clip is anomalous under weak temporal supervision. This formulation makes the deployment challenge explicit: it forces the model to commit to an actionable decision boundary rather than only ranking clips by anomaly likelihood. We conduct a systematic evaluation across the ShanghaiTech and CHAD datasets, varying prompt specificity and clip duration to characterize how MLLMs behave under limited temporal evidence and noisy surveillance context, and to identify practical prompt design factors that consistently shape anomaly decisions. Our evaluation achieved a peak F1-score of 0.64 on the ShanghaiTech dataset\cite{shanghaitech}, whereas the increased complexity of the CHAD dataset \cite{chad} limited performance to a maximum F1 of 0.48, underscoring the significant challenge of high-resolution real-world anomaly detection.

This study makes the following contributions:
\begin{itemize}
\item \textbf{A deployment-oriented VAD formulation for MLLMs:} We cast anomaly detection as a prompt-guided binary decision problem under weak temporal supervision, explicitly targeting the real-world \emph{decision boundary} requirement rather than only offline ranking behavior.
\item \textbf{A controlled analysis of prompting as a decision interface:} We study prompt design along the axis of \emph{specificity vs. verbosity} (e.g., concise vs. detailed instructions; generic vs. class-aware guidance) to understand how language constraints shape MLLM anomaly judgments in surveillance footage.
\item \textbf{Temporal-context probing for surveillance realism:} We evaluate how short clip windows (1s/2s/3s) modulate model sensitivity and stability, and we provide practical guidelines for configuring temporal context when applying MLLMs to VAD.
\end{itemize}

\section{Related Works}

Video Anomaly Detection (VAD) is a fundamental problem in computer vision that aims to automatically identify unusual events or behaviors in video sequences. It plays a critical role in applications such as smart surveillance \cite {noghre2024human, rashvand2025shopformer}, healthcare monitoring \cite{nanda2023soft}, autonomous driving (accident detection) and traffic monitoring \cite{zhao2023unsupervised,11063431, yu2022deep,10794646}. Among its subdomains, human-centric anomaly detection has received increasing attention, focusing specifically on identifying atypical human behaviors \cite{yao2026offline, Noghre_2024_WACV,Rashvand_2025_WACV}. This setting is particularly challenging due to the rare events occurring infrequently in real-world data, resulting in limited representative samples for supervised training.
Early VAD systems relied on handcrafted features \cite{cheng2015gaussian, cocsar2016toward}, such as optical flow statistics and trajectory descriptors, which struggled to generalize to complex real-world environments. The emergence of deep learning has introduced diverse learning and architectural paradigms for VAD. Most existing VAD frameworks adopt semi-supervised or unsupervised learning paradigms \cite{tian2021weakly, li2023essl}, where models are trained exclusively on normal behavior and anomalies are defined as deviations from learned distributions. These approaches typically rely on reconstruction-based objectives, future or past prediction strategies, or multi-branch architectures to model normal motion patterns \cite{noghre2025survey}. Within this learning setting, diverse architectural designs have been developed to better capture spatio-temporal dynamics, including Autoencoders for self-supervised feature learning \cite{Noghre_2024_WACV}, Long Short-Term Memory (LSTM) models, and Spatio-Temporal Graph Convolutional Networks (ST-GCNs) for temporal and graph-based motion modeling \cite{noghre2025survey}, and Transformer architectures \cite{noghre2024human, rashvand2025shopformer,vaswani2017attention} for capturing complex and long-context dependencies. Modern VAD algorithms are broadly categorized into pixel-based \cite{kirichenko2022detection, martinez2021criminal} and pose-based \cite{hirschorn2023normalizing, yu2023regularity} approaches. Pixel-based models operate directly on raw RGB frames to learn appearance and motion patterns, often achieving strong performance but remaining sensitive to background variations and illumination changes \cite{noghre2025survey,kirichenko2022detection, martinez2021criminal}. In contrast, pose-based VAD abstracts human activity into skeletal representations, focusing solely on motion dynamics while discarding identifiable visual information. This abstraction improves robustness to scene variability and provides a privacy-aware alternative for surveillance applications \cite{noghre2025survey,hirschorn2023normalizing, yu2023regularity,markovitz2020graph}.

\begin{figure*}[h]
    \centering
    \includegraphics[clip,trim={0 80 0 0},width=1\textwidth]{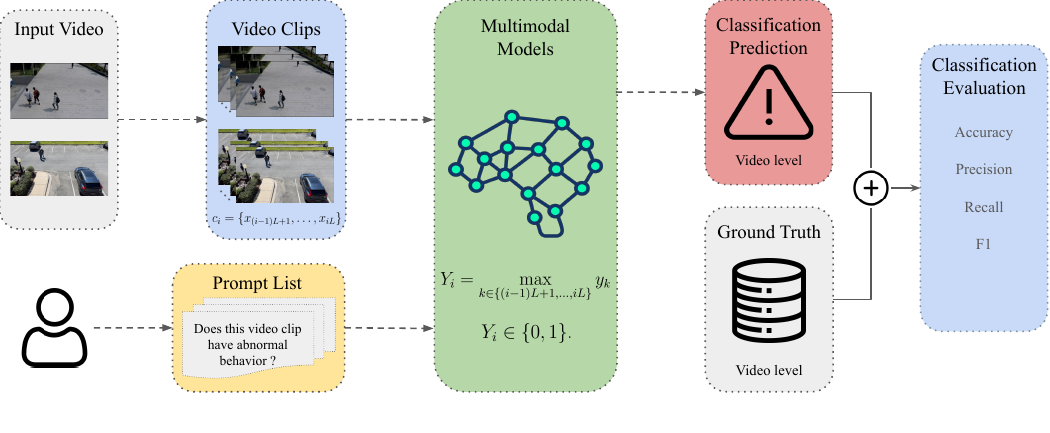}
    \caption{System architecture for prompt-based video anomaly detection. The workflow illustrates the transformation of raw input videos into segmented video clips, which are then processed by Multimodal Models guided by different prompts. This pipeline yields a classification prediction that is directly evaluated against the established ground truth at the Video level}
    \label{fig:dataflow}
\end{figure*}

The growing adoption of transformers in VAD parallels the broader success of Large Language Models (LLMs) in sequence modeling and reasoning \cite{minaee2024large,kalyan2024survey, 10903265, wang2024survey,ghorbani2025examining}. Originally developed for natural language processing, LLMs have evolved into MLLMs capable of integrating textual, visual, and temporal representations within unified architectures \cite{zhang2024mm,yin2024survey,li2025systematic}. 
In parallel, video generation has experienced remarkable advancements. State-of-the-art frameworks such as Sora \cite{he2025pre}, VideoPoet \cite{kondratyuk2023videopoet}, and Make-A-Video \cite{singer2022make} demonstrate high-fidelity, temporally coherent video synthesis from textual prompts. Foundation models such as Gemini \cite{team2024gemini} and GPT-4o \cite{hurst2024gpt} further extend this ecosystem by enabling unified reasoning across text, images, audio, and video. The rapid development of AI video creators has motivated new evaluation benchmarks, including Video-Bench \cite{han2025video} and VBench-2.0   \cite{zheng2025vbench}, which assess visual quality, physical consistency, and prompt alignment in generated videos.

While current MLLMs can produce visually complex and temporally consistent sequences, their ability to deeply understand and operationally interpret dynamic events remains largely underexplored. Existing evaluation frameworks primarily focus on generative fidelity or guided reasoning rather than operational performance. Benchmarks such as Video-MME \cite{fu2025video}, TempCompass \cite{liu2024tempcompass} and UCVL \cite{chen2025ucvl} assess video understanding through structured multiple-choice or question-answering formats. Although these benchmarks provide valuable insights into reasoning and alignment capabilities, they rely on passive evaluation with predefined answer choices. Such settings do not fully evaluate the autonomous and real-time video understanding required for real-world video reasoning and surveillance systems.\cite{wu2024vadclip,zanella2024delving} Motivated by this gap and the critical importance of VAD for public safety, 
our work directly advances beyond descriptive or guided evaluation. Rather than measuring whether MLLMs can narrate or critique video content, we assess their ability to function as autonomous anomaly detectors under realistic deployment conditions. By shifting the emphasis from generative storytelling and guided question-answering to continuous, real-time detection reliability, our framework evaluates a deeper level of video understanding, one that reflects operational deployment requirements and remains largely unexplored by existing generative and reasoning benchmarks.

\section{Methodology}
\label{sec:method}

\subsection{Problem Setup and Notation}
We formulate \emph{video anomaly detection} as a binary classification problem under weak temporal supervision. A video is represented as an ordered sequence of frames $V=\{x_1,x_2,\dots,x_T\}$, where each frame $x_t \in \mathbb{R}^{H \times W \times C}$. Frame-level ground-truth labels are denoted by $y_t \in \{0,1\}$, with $y_t=1$ indicating an anomalous frame. Since MLLMs such as Gemini operate on short temporal windows, we convert each test video into fixed-length clips and evaluate Gemini's ability to classify each clip as \emph{normal} or \emph{anomalous}.

\begin{table*}[]
\centering
\caption{Summary of Top 3 Performing Prompt Configurations by Peak F1-Score}
\label{tab:prompt}
\resizebox{0.95\textwidth}{!}{%
\begin{tabular}{c|ll|c}
\toprule
\toprule
\textbf{Rank} & \textbf{Prompt Base} & \textbf{} & \textbf{Peak F1 (with class)} \\ \midrule
1 & \begin{tabular}[c]{@{}l@{}}GPT instant \\ medium\end{tabular} & \begin{tabular}[c]{@{}l@{}}Analyze the provided one-second video clip and decide whether it \\ contains any anomalous human-related behavior. An anomalous \\ behavior is any human action or event that is out of the ordinary,\\  unexpected, unsafe, or inconsistent with normal real-world human \\ activities and environments. Routine, everyday human actions should \\ be considered non-anomalous. Based only on the visual content, \\ output a binary decision. Output 1 if an anomaly is present. \\ Output 0 if no anomaly is present. Do not include any other text.\end{tabular} & 0.64 \\ \midrule
2 & \begin{tabular}[c]{@{}l@{}}Gemini fast \\ medium\end{tabular} & \begin{tabular}[c]{@{}l@{}}Analyze this one-second video clip to determine if it contains \\ any anomalous human behavior. For the purposes of this evaluation,\\  an anomaly includes any human action or interaction that is \\ unusual, unsafe, or unexpected in a typical real-world environment. \\ Ordinary daily activities should be considered normal. Examine the \\ clip for any events that should not normally occur and decide if the \\ behavior shown is anomalous or not anomalous. You must indicate \\ your decision by outputting a binary label where 1 represents an \\ anomaly and 0 represents a normal event. Do not include any \\ descriptions or additional commentary in your response. \\ The only acceptable output is the single digit 1 or 0.\end{tabular} & 0.63 \\ \midrule
3 & \begin{tabular}[c]{@{}l@{}}Gemini pro \\ short\end{tabular} & \begin{tabular}[c]{@{}l@{}}Examine this one-second video for anomalous human \\ behavior which is defined as unexpected or unsafe action \\ distinct from normal daily activities. Output 1 if an anomaly\\  is present and 0 if the behavior is normal. Provide only the \\ binary digit as your answer with no other text.\end{tabular} & 0.59 \\ \midrule
4 & \begin{tabular}[c]{@{}l@{}}Gemini pro \\ long\end{tabular} & \begin{tabular}[c]{@{}l@{}}You are an advanced video analysis system tasked with \\ evaluating a one-second video clip to detect anomalous \\ human behavior. For the purpose of this analysis, an anomaly\\  is strictly defined as any human action, interaction, or event \\ that is out of the ordinary, unexpected, unsafe, or deviates from \\ standard real-world norms and environmental contexts. Normal,\\  routine daily human activities are not considered anomalies and \\ should be classified as negative. You must rigorously analyze the \\ visual content to distinguish between standard behavior and \\ irregular occurrences. Your output must be a strict binary \\ classification. If the video contains an anomaly, output the digit 1.\\  If the video contains only normal behavior, output the digit 0. \\ You are prohibited from providing any text, explanation, \\ reasoning, or justification.\end{tabular} & 0.58 \\ \midrule
5 & \begin{tabular}[c]{@{}l@{}}GPT instant \\ short\end{tabular} & \begin{tabular}[c]{@{}l@{}}Watch the one-second video clip and determine \\ whether any anomalous human-related behavior is \\ present. Output only 1 for anomalous or 0 for not anomalous.\end{tabular} & 0.57 \\
\bottomrule
\bottomrule
\end{tabular}%
}

\end{table*}

\subsection{Datasets}
We evaluate on the \textbf{test splits} of two widely used benchmarks for anomaly detection in surveillance video: CHAD and ShanghaiTech. Both datasets provide frame-level annotations of normal versus anomalous events, which allows us to derive clip-level labels and perform quantitative evaluation under a consistent clip-based protocol.

\subsection{Data Preparation and Clip Construction}
In real-world surveillance, the priority is the early detection of an event's onset, such as the first strike in a fight or the initial loss of balance in a fall.\cite{rashvand2026frames} We restrict our analysis to $t \in \{1,2,3\}$ seconds to prioritize low-latency response, as longer temporal windows—while providing more context, inherently delay the triggering of time, sensitive alerts in autonomous security systems. 

Each test video is segmented into non-overlapping clips of duration $t \in \{1,2,3\}$ seconds. Given a frame rate of $r$ frames per second (FPS), each clip contains $L=t\cdot r$ frames. For a video with $T$ frames, this yields $N=\left\lfloor T/L \right\rfloor$ clips, where the $i$-th clip is
\begin{equation}
c_i=\{x_{(i-1)L+1},\dots,x_{iL}\}, \qquad i=1,\dots,N .
\end{equation}
We propagate frame-level labels to clip-level labels using an \emph{any-anomaly} rule: a clip is considered anomalous if it contains at least one anomalous frame. Concretely, the clip label is defined as
\begin{equation}
Y_i=\max_{k \in \{(i-1)L+1,\dots,iL\}} y_k, \qquad Y_i \in \{0,1\}.
\end{equation}
Thus, $Y_i=1$ if any frame in $c_i$ is anomalous and $Y_i=0$ otherwise.

\begin{table*}[htp]
\centering
\caption{Performance evaluation of various configurations on the ShanghaiTech dataset\cite{shanghaitech}. The results compare F1-score, Precision, and Recall across three temporal clip lengths (1s, 2s, and 3s) and evaluate the impact of including class-specific instructions in the prompt list.}
\label{tab:sht_all}
\resizebox{0.95\textwidth}{!}{
\begin{tabular}{l|ccc|ccc|ccc}
\toprule
\toprule
\textbf{Prompt} & \multicolumn{3}{c|}{\textbf{F1-score}} & \multicolumn{3}{c|}{\textbf{Precision}} & \multicolumn{3}{c}{\textbf{Recall}} \\ \midrule
 & \textbf{1s} & \textbf{2s} & \textbf{3s} & \textbf{1s} & \textbf{2s} & \textbf{3s} & \textbf{1s} & \textbf{2s} & \textbf{3s} \\ \midrule
Human & 0.04 & 0.09 & 0.10 & \textbf{100\%} & \textbf{100\%} & \underline{94.12\%} & 1.85\% & 4.91\% & 5.08\% \\
Human + class & 0.39 & 0.45 & 0.51 & 79.92\% & 88.39\% & 88.89\% & 26.08\% & 30.58\% & 35.67\% \\
\midrule
GPT instant long & 0.11 & 0.04 & 0.06 & 84.48\% & \textbf{100\%} & \textbf{100\%} & 6.05\% & 2.23\% & 2.86\% \\
GPT instant long + class & 0.13 & 0.14 & 0.16 & 85.07\% & \textbf{100\%} & 90.32\% & 7.06\% & 7.59\% & 8.89\% \\
\midrule
GPT instant medium & 0.06 & 0.06 & 0.09 & 89.66\% & \textbf{100\%} & \textbf{100\%} & 3.23\% & 3.12\% & 4.46\% \\
GPT instant medium + class & 0.49 & \textbf{0.59} & \textbf{0.64} & 78.71\% & 81.25\% & 81.16\% & 36.05\% & \textbf{46.53\%} & \textbf{53.33\%} \\
\midrule
GPT instant short & 0.06 & 0.09 & 0.16 & 90.00\% & \textbf{100\%} & 93.33\% & 3.33\% & 4.94\% & 8.89\% \\
GPT instant short + class & 0.46 & 0.52 & 0.57 & 81.42\% & 82.39\% & 85.07\% & 32.43\% & 38.84\% & 43.49\% \\
\midrule
GPT think long & 0.03 & 0.04 & 0.12 & \textbf{100\%} & \textbf{100\%} & \textbf{100\%} & 1.60\% & 1.84\% & 6.17\% \\
GPT think long + class & 0.20 & 0.23 & 0.28 & 83.19\% & 90.91\% & 86.84\% & 11.59\% & 12.28\% & 18.02\% \\
\midrule
GPT think medium & 0.01 & 0.03 & 0.01 & \textbf{100\%} & \textbf{100\%} & \textbf{100\%} & 0.74\% & 1.37\% & 0.69\% \\
GPT think medium + class & 0.30 & 0.38 & 0.47 & 81.52\% & 82.65\% & 90.54\% & 18.61\% & 25.45\% & 32.29\% \\
\midrule
GPT think short & 0.07 & 0.06 & 0.11 & 84.85\% & \underline{94.12\%} & \textbf{100\%} & 3.45\% & 2.90\% & 6.02\% \\
GPT think short + class & 0.45 & 0.49 & 0.54 & 82.30\% & 89.25\% & 87.98\% & 31.06\% & 33.71\% & 39.21\% \\
\midrule
Gemini fast long & 0.03 & 0.05 & 0.06 & \textbf{100\%} & \textbf{100\%} & \textbf{100\%} & 1.48\% & 2.45\% & 2.86\% \\
Gemini fast long + class & 0.23 & 0.31 & 0.30 & 84.92\% & 88.68\% & 79.44\% & 13.21\% & 19.20\% & 19.21\% \\
\midrule
Gemini fast medium & 0.09 & 0.10 & 0.10 & 90.48\% & 90.62\% & 87.10\% & 4.69\% & 5.13\% & 5.02\% \\
Gemini fast medium + class & \textbf{0.52} & 0.55 & \underline{0.63} & 77.45\% & 85.62\% & 87.58\% & \textbf{39.60\%} & 40.85\% & \underline{48.06\%} \\
\midrule
Gemini fast short & 0.10 & 0.11 & 0.12 & 87.23\% & \textbf{100\%} & \textbf{100\%} & 5.06\% & 5.80\% & 6.50\% \\
Gemini fast short + class & 0.31 & 0.36 & 0.43 & 82.80\% & 83.21\% & 89.15\% & 18.99\% & 23.66\% & 29.31\% \\
\midrule
Gemini pro long & 0.05 & 0.05 & 0.03 & \underline{91.67\%} & \textbf{100\%} & \textbf{100\%} & 2.80\% & 2.45\% & 1.37\% \\
Gemini pro long + class & 0.47 & 0.51 & 0.58 & 80.00\% & 87.91\% & 87.70\% & 33.00\% & 36.83\% & 43.10\% \\
\midrule
Gemini pro medium & 0.02 & 0.06 & 0.04 & \textbf{100\%} & \textbf{100\%} & \textbf{100\%} & 1.23\% & 2.96\% & 2.08\% \\
Gemini pro medium + class & 0.30 & 0.41 & 0.47 & 80.75\% & 86.48\% & 86.43\% & 18.69\% & 28.12\% & 33.07\% \\
\midrule
Gemini pro short & 0.13 & 0.27 & 0.29 & 84.62\% & 92.11\% & 82.35\% & 6.81\% & 15.70\% & 17.78\% \\
Gemini pro short + class & \underline{0.51} & \underline{0.59} & 0.60 & 77.97\% & 81.35\% & 80.75\% & \underline{38.12\%} & \underline{45.86\%} & 47.94\% \\
\bottomrule
\bottomrule
\end{tabular}
}
\end{table*}

\subsection{Prompt Generation}
We study prompt specificity as a controlled experimental factor by evaluating Gemini 2.5 Flash Lite, the sole inference engine, under a set of four prompts. The first is a human-authored base prompt, denoted $p^{(0)}$. The remaining three prompts $\{p^{(1)},p^{(2)},p^{(3)}\}$ are produced by prompting multiple LLM variants namely \textbf{ChatGPT 5.2 Instant}, \textbf{ChatGPT 5.2 Thinking}, \textbf{Gemini 3 Fast}, \textbf{Gemini 3 Thinking}, and \textbf{Gemini 3 Pro} to generate candidate prompts, from which we construct the final three specificity-level prompts used in evaluation. All models are given the same human-written meta-prompt to ensure a consistent generation objective.

The meta-prompt instructs the model to generate three semantically equivalent prompts for classifying one-second video clips as anomalous versus non-anomalous, differing only in the degree of detail. It defines an anomaly as any human action, interaction, or event that is out of the ordinary, unexpected, unsafe, or inappropriate under common real-world conditions, and specifies a strict binary output constraint: the model must output \texttt{1} (Anomalous) or \texttt{0} (Not Anomalous) with no additional text. The three generated prompts correspond to a \emph{comprehensive specification} version (maximally explicit definitions, assumptions, and criteria), a \emph{structured instruction} version (moderate detail with clear task framing), and a \emph{minimal instruction} version (concise task-only guidance). We denote the full prompt set as $\mathcal{P}=\{p^{(0)},p^{(1)},p^{(2)},p^{(3)}\}$ and query Gemini with each $p^{(m)} \in \mathcal{P}$ for every clip to measure how instruction granularity affects anomaly recognition and decision consistency.

To evaluate the impact of domain-specific knowledge, we introduce a Class-Aware (CA) variant for each prompt. The only difference between the standard prompt and its CA counterpart is the inclusion of the specific anomaly labels provided by the original dataset authors (e.g., 'fighting,' 'theft'). For example, the CA instruction appends the following string: 'The anomalies to detect include: [List of classes from dataset paper].' Results labeled 'With Class' in \cref{tab:sht_all} and \cref{tab:chad_all} refer to this configuration.

\begin{table*}[htp]
\centering
\caption{Performance evaluation of various configurations on the CHAD dataset \cite{chad} The results compare F1-score, Precision, and Recall across three temporal clip lengths (1s, 2s, and 3s) and evaluate the impact of including class-specific instructions in the prompt list.}
\label{tab:chad_all}
\resizebox{0.95\textwidth}{!}{
\begin{tabular}{l|ccc|ccc|ccc}
\toprule
\toprule
\textbf{Prompt} & \multicolumn{3}{c|}{\textbf{F1-score}} & \multicolumn{3}{c|}{\textbf{Precision}} & \multicolumn{3}{c}{\textbf{Recall}} \\ \midrule
 & \textbf{1s} & \textbf{2s} & \textbf{3s} & \textbf{1s} & \textbf{2s} & \textbf{3s} & \textbf{1s} & \textbf{2s} & \textbf{3s} \\ \midrule
Human & 0.13 & 0.14 & 0.14 & \underline{96.82\%} & \textbf{97.75\%} & \textbf{98.46\%} & 7.07\% & 7.51\% & 7.76\% \\
Human + class & 0.23 & 0.17 & 0.12 & 93.56\% & 93.16\% & 90.16\% & 12.92\% & 9.42\% & 6.66\% \\
\midrule
GPT instant long & 0.15 & 0.13 & 0.11 & 92.31\% & 93.18\% & 92.59\% & 8.36\% & 7.09\% & 6.05\% \\
GPT instant long + class & 0.23 & 0.24 & 0.23 & 90.26\% & 90.06\% & 90.08\% & 13.45\% & 14.12\% & 13.21\% \\
\midrule
GPT instant medium & 0.13 & 0.17 & 0.13 & \textbf{97.44\%} & 94.59\% & 95.10\% & 7.06\% & 9.07\% & 7.18\% \\
GPT instant medium + class & 0.21 & 0.21 & 0.20 & 90.91\% & 92.52\% & 89.34\% & 12.26\% & 12.59\% & 11.38\% \\
\midrule
GPT instant short & 0.25 & 0.23 & 0.18 & 94.20\% & 95.92\% & 93.67\% & 14.68\% & 13.55\% & 9.88\% \\
GPT instant short + class & 0.30 & 0.28 & 0.25 & 90.43\% & 91.00\% & 91.52\% & 18.54\% & 16.76\% & 14.53\% \\
\midrule
GPT think long & 0.11 & 0.06 & 0.04 & 89.74\% & 90.91\% & 100\% & 5.65\% & 3.07\% & 1.94\% \\
GPT think long + class & 0.24 & 0.20 & 0.16 & 86.67\% & 90.52\% & 87.78\% & 14.27\% & 11.32\% & 9.12\% \\
\midrule
GPT think medium & 0.17 & 0.12 & 0.08 & 93.48\% & 94.74\% & 96.88\% & 9.69\% & 6.42\% & 4.15\% \\
GPT think medium + class & 0.32 & 0.30 & 0.25 & 88.30\% & 90.09\% & 89.76\% & 20.28\% & 18.24\% & 14.41\% \\
\midrule
GPT think short & 0.22 & 0.18 & 0.12 & 90.00\% & 93.33\% & 93.75\% & 12.49\% & 9.83\% & 6.58\% \\
GPT think short + class & 0.37 & 0.35 & 0.29 & 89.19\% & 89.04\% & 90.50\% & 24.36\% & 22.41\% & 17.30\% \\
\midrule
Gemini fast long & 0.05 & 0.06 & 0.04 & 90.00\% & 93.75\% & 92.86\% & 2.57\% & 3.12\% & 2.08\% \\
Gemini fast long + class & 0.29 & 0.28 & 0.24 & 87.35\% & 88.59\% & 90.21\% & 18.49\% & 16.69\% & 13.70\% \\
\midrule
Gemini fast medium & 0.18 & 0.19 & 0.16 & 94.55\% & 95.65\% & 95.65\% & 10.32\% & 10.86\% & 8.80\% \\
Gemini fast medium + class & \textbf{0.44} & 0.44 & 0.38 & 89.87\% & 90.32\% & 91.74\% & \textbf{30.51\%} & 29.94\% & 24.42\% \\
\midrule
Gemini fast short & 0.20 & 0.20 & 0.15 & 94.78\% & 95.71\% & 96.61\% & 11.39\% & 11.65\% & 7.94\% \\
Gemini fast short + class & 0.39 & 0.40 & 0.35 & 89.67\% & 90.49\% & 91.53\% & 26.29\% & 26.46\% & 21.95\% \\
\midrule
Gemini pro long & 0.04 & 0.04 & 0.03 & 93.33\% & 92.86\% & 94.44\% & 2.04\% & 2.13\% & 1.51\% \\
Gemini pro long + class & \underline{0.42} & \textbf{0.48} & \underline{0.45} & 89.16\% & 90.23\% & 91.91\% & \underline{28.33\%} & \textbf{33.59\%} & \underline{30.63\%} \\
\midrule
Gemini pro medium & 0.09 & 0.12 & 0.12 & 92.59\% & \underline{97.50\%} & \underline{98.00\%} & 4.94\% & 6.38\% & 6.32\% \\
Gemini pro medium + class & 0.33 & 0.33 & 0.29 & 87.68\% & 89.08\% & 91.36\% & 21.07\% & 21.03\% & 17.19\% \\
\midrule
Gemini pro short & 0.27 & 0.26 & 0.22 & 92.66\% & 92.88\% & 94.20\% & 16.25\% & 14.99\% & 12.17\% \\
Gemini pro short + class & 0.40 & \underline{0.47} & \textbf{0.46} & 85.89\% & 85.45\% & 90.78\% & 25.77\% & \underline{32.47\%} & \textbf{30.99\%} \\
\bottomrule
\bottomrule
\end{tabular}
}
\end{table*}

\subsection{Gemini-Based Video Anomaly Inference}
For each clip $c_i$ and prompt $p^{(m)}$, we submit the pair $(c_i,p^{(m)})$ to Gemini Video Understanding and obtain a textual response $R_i^{(m)}$. We then convert the response into a binary prediction using a deterministic parsing function $g(\cdot)$, yielding $\hat{Y}_i^{(m)}=g(R_i^{(m)})$ with $\hat{Y}_i^{(m)} \in \{0,1\}$. In practice, $g(\cdot)$ maps the response to \emph{anomalous} versus \emph{normal} based on the required output format and the presence of explicit anomaly-related statements. Querying multiple prompts produces prompt-conditioned predictions $\{\hat{Y}_i^{(m)}\}_{m=0}^{3}$ for the same clip, enabling a direct comparison of performance and robustness across prompt specificity levels.

\section{Experiment and Evaluation}

The experimental framework utilizes a multimodal reasoning approach to classify anomaly within video data shown in \cref{fig:dataflow}. The process initiates with the pre-process of raw input video stream, which is divded into discrete video clips to have better temporal analysis. These clips, ranging from one second to three seconds, serve as the visual component for the Multimodal Models, while a user, issued prompt list, containing queries such as "Does this video clip have abnormal behavior?"; provides the necessary textual context for zero-shot or directed reasoning. Upon processing these dual inputs, the model generates a classification prediction for each input clip, categorized at the video level. To quantify the model's performance, these predictions are aggregated or compared against the processed ground truth labels for each corresponding video segment to determine detection accuracy.

To ensure the integrity and fairness of our comparative evaluation, Gemini 2.5 Flash Lite was selected as the primary MLLM for the Zero-Shot VAD task. While contemporary models such as GPT-5 claim video understanding capabilities, their architectures often rely on external preprocessing that decompose video into a sequence of discrete image frames and audio tracks, thereby losing temporal continuity and motion-based context. In contrast, Gemini 2.5 offers native video analysis without the need for manual frame-level preprocessing or 'filmstrip' conversions. This architectural advantage allows for superior spatio-temporal reasoning, making Gemini the only suitable candidate for a truly zero-shot evaluation where the model must interpret fluid motion and temporal irregularities directly from the raw video stream.

As mentioned in \cref{sec:method}, an analysis of the experimental results across both datasets reveals a significant disparity between precision and recall, as shown in \cref{tab:sht_all} and \cref{tab:chad_all}. In the proposed workflow, multimodal models process video clips to generate a classification prediction at the video level. Across nearly all base configurations without class-specific augmentation, the models exhibit an extreme "conservative bias." For instance, in the SHT dataset (\cref{tab:sht_all}), several generated prompt base such as GPT think medium and Gemini pro medium achieved a precision of 100\%, yet their recall frequently fell below 3\%. This indicates that while the models are highly accurate when they do trigger an anomaly flag, they are fundamentally biased toward a "no-anomaly" baseline, failing to detect the vast majority of abnormal behaviors. This suggests that current MLLMs, despite their general-purpose reasoning capabilities, struggle with the high-stakes sensitivity required for real-world security applications where missing an event is as critical as a false positive.

The comparative effectiveness of the various prompt setups is summarized in \cref{tab:prompt}, which identifies the top five performing configurations based on their peak F1-scores. Notably, the GPT instant medium + class configuration leads the results with an F1-score of 0.64 on the ShanghaiTech dataset, followed closely by Gemini fast medium + class. The inclusion of this table serves to consolidate the findings, demonstrating that while different models excel on different datasets, the combination of medium-to-short prompt lengths with explicit categorical context remains the most robust strategy for maximizing detection accuracy in multimodal video analysis.
\begin{table}[]
\centering
\caption{Comparative analysis of the change in F1-score ($\Delta F_1 = F_{1,\text{class}} - F_{1,\text{no-class}}$) as adding anomalous class to the prompt bases. This table quantifies the performance gains (or losses) when transitioning from non-class to with-class across both the ShanghaiTech and CHAD datasets.}
\label{tab:class}
\resizebox{\columnwidth}{!}{%
\begin{tabular}{c|l|cccc}
\toprule
\toprule
\textbf{Dataset} & \textbf{Prompt Base} & \textbf{1s $\Delta$F1} & \textbf{2s $\Delta$F1} & \textbf{3s $\Delta$F1} & \textbf{Mean $\Delta$F1} \\ \hline
\multirow{13}{*}{\begin{tabular}[c]{@{}c@{}}ShanghaiTech\\ \cite{shanghaitech}\end{tabular}} & Human & +0.36 & +0.36 & +0.41 & +0.38 \\
 & GPT instant long & +0.02 & +0.10 & +0.11 & +0.07 \\
 & GPT instant medium & +0.43 & +0.53 & +0.56 & +0.51 \\
 & GPT instant short & +0.40 & +0.42 & +0.41 & +0.41 \\
 & GPT think long & +0.17 & +0.19 & +0.17 & +0.18 \\
 & GPT think medium & +0.29 & +0.36 & +0.46 & +0.37 \\
 & GPT think short & +0.38 & +0.43 & +0.43 & +0.41 \\
 & Gemini fast long & +0.20 & +0.27 & +0.25 & +0.24 \\
 & Gemini fast medium & +0.43 & +0.45 & +0.53 & \textbf{+0.47} \\
 & Gemini fast short & +0.21 & +0.25 & +0.31 & +0.26 \\
 & Gemini pro long & +0.41 & +0.47 & +0.55 & +0.48 \\
 & Gemini pro medium & +0.28 & +0.35 & +0.43 & +0.36 \\
 & Gemini pro short & +0.39 & +0.32 & +0.31 & +0.34 \\ \midrule
\multirow{13}{*}{\begin{tabular}[c]{@{}c@{}}CHAD\\ \cite{chad}\end{tabular}} & Human & +0.10 & +0.03 & -0.02 & +0.04 \\
 & GPT instant long & +0.08 & +0.11 & +0.12 & +0.10 \\
 & GPT instant medium & +0.08 & +0.04 & +0.07 & +0.06 \\
 & GPT instant short & +0.05 & +0.05 & +0.07 & +0.06 \\
 & GPT think long & +0.13 & +0.14 & +0.12 & +0.13 \\
 & GPT think medium & +0.15 & +0.18 & +0.17 & +0.17 \\
 & GPT think short & +0.15 & +0.17 & +0.17 & +0.16 \\
 & Gemini fast long & +0.24 & +0.22 & +0.20 & +0.22 \\
 & Gemini fast medium & +0.26 & +0.25 & +0.22 & +0.24 \\
 & Gemini fast short & +0.19 & +0.20 & +0.20 & +0.20 \\
 & Gemini pro long & +0.38 & +0.44 & +0.42 & \textbf{+0.41} \\
 & Gemini pro medium & +0.24 & +0.21 & +0.17 & +0.21 \\
 & Gemini pro short & +0.13 & +0.21 & +0.24 & +0.19 \\ 
\bottomrule
\bottomrule
\end{tabular}%
}
\end{table}

The results further challenge the assumption that increased prompt detail leads to superior model performance. As illustrated in the experimental flow, the models utilize a prompt list to evaluate whether a video clip contains abnormal behavior. However, comparing the "short," "medium," and "long" prompt results in \cref{tab:sht_all,tab:chad_all} suggests that longer, more detailed instructions do not consistently improve the F1-score. In many cases, "medium" prompts outperformed "long" ones, suggesting that excessive detail may introduce semantic noise that distracts the reasoning engine. The most significant performance catalyst was the inclusion of class-specific context (the "+ class" rows). Adding this specific context dramatically improved the recall, often by a factor of five or more, as seen in the Gemini fast medium configuration in \cref{tab:sht_all}, where recall jumped from 4.69\% to 39.60\%. This indicates that the bottleneck in MLLM-based video anomaly detection is not necessarily a lack of visual recognition, but rather a lack of categorical confidence that is only resolved through explicit prompt-level guidance.
\begin{table}[]
\centering
\caption{Comparative analysis of the change in F1-score ($\Delta$) as temporal clip length increases. This table quantifies the performance gains (or losses) when transitioning from a 1s window to 2s and 3s windows across both the ShanghaiTech and CHAD datasets.}
\label{tab:time}
\resizebox{\columnwidth}{!}{%
\begin{tabular}{c|l|cc}
\toprule
\toprule
\textbf{Dataset} & \textbf{Prompt} & \textbf{$\Delta(1s\rightarrow 2s)$} & \textbf{$\Delta(1s\rightarrow 3s)$} \\ \hline
\multirow{13}{*}{\begin{tabular}[c]{@{}c@{}}ShanghaiTech\\ \cite{shanghaitech}\end{tabular}} & Human & +0.06 & +0.09 \\
 & GPT instant long & -0.03 & -0.01 \\
 & GPT instant medium & +0.05 & +0.09 \\
 & GPT instant short & +0.07 & +0.08 \\
 & GPT think long & +0.06 & +0.09 \\
 & GPT think medium & +0.15 & +0.13 \\
 & GPT think short & +0.06 & +0.11 \\
 & Gemini fast long & -0.01 & +0.05 \\
 & Gemini fast medium & +0.03 & +0.05 \\
 & Gemini fast short & +0.04 & +0.10 \\
 & Gemini pro long & +0.11 & +0.17 \\
 & Gemini pro medium & -0.02 & -0.03 \\
 & Gemini pro short & +0.11 & +0.13 \\ \midrule
\multirow{13}{*}{\begin{tabular}[c]{@{}c@{}}CHAD\\ \cite{chad}\end{tabular}} & Human & -0.02 & -0.05 \\
 & GPT instant long & -0.01 & -0.02 \\
 & GPT instant medium & +0.02 & -0.01 \\
 & GPT instant short & +0.02 & +0.01 \\
 & GPT think long & +0.00 & -0.01 \\
 & GPT think medium & +0.08 & +0.05 \\
 & GPT think short & +0.04 & +0.05 \\
 & Gemini fast long & -0.03 & -0.04 \\
 & Gemini fast medium & +0.01 & -0.03 \\
 & Gemini fast short & +0.00 & -0.00 \\
 & Gemini pro long & +0.04 & +0.02 \\
 & Gemini pro medium & -0.03 & -0.05 \\
 & Gemini pro short & +0.05 & +0.07 \\ 
\bottomrule
\bottomrule
\end{tabular}%
}
\end{table}

Despite the technical improvements of the CHAD dataset, which features higher resolution and frame rates compared to SHT, there was no significant leap in model performance. As shown in \cref{tab:chad_all}, the peak F1-scores remained lower than those observed in the SHT experiments, with Gemini pro long + class reaching only 0.48 compared to the 0.64 achieved by GPT on SHT. This observation is critical: it suggests that higher visual fidelity does not solve the underlying challenges of video understanding in anomaly detection. Even though the input video is of higher quality, the model's ability to interpret the scene against the ground truth  remains limited. These results demonstrate that while current MLLMs serve as a strong baseline for general video understanding, they still fall short when transitioned to real-world, high-resolution applications that require more behavioral interpretation rather than simple object recognition.

A critical finding illustrated in \cref{tab:class} is the transformative effect of adding class instructions to the prompt list. Across all evaluated models, the base configuration consistently suffered from a severe precision-recall imbalance, where the model was highly precise but missed nearly all relevant events. By appending class-specific context, the "+ class" rows, we observe a dramatic "Recall-Correction" effect. For instance, Gemini pro short on SHT saw its F1-score jump from 0.13 to 0.51 at the $1s$ interval, driven by a recall increase from 6.81\% to 38.12\%. This shift proves that the MLLM's primary failure mode is not a lack of visual "sight," but a lack of "intent." Without specific classes to monitor, the model defaults to a conservative non-trigger state; with them, it gains the categorical confidence required to identify anomalies in a real-world scenario.

The influence of clip length on model performance, detailed in \cref{tab:time}, suggests that longer temporal windows generally assist the reasoning capabilities of MLLMs, though the effect is not significant and  varies significantly by dataset. In the ShanghaiTech dataset, most models exhibited a positive F1-score delta when increasing the clip length from $1s$ to $3s$. For example, Gemini pro long saw a substantial improvement of +0.17, while GPT think medium improved by +0.13. This trend suggests that for lower-resolution datasets like SHT, additional temporal context is vital for the model to distinguish between normal and abnormal movements. Conversely, the CHAD dataset showed much more marginal improvements or even performance degradation, such as the -0.05 drop for Gemini pro medium. This indicates that in higher-resolution environments, simply increasing the temporal window does not resolve the underlying semantic confusion of the model, and in some cases, may introduce redundant information that obscures the anomaly.

\section{Conclusion}

The experimental evaluation of multimodal Large Language Models (MLLMs) for real-world video anomaly detection exposes a clear gap between general video understanding and dependable operational behavior. While our framework demonstrates that MLLMs can be integrated into a surveillance-style pipeline, consuming raw clips and producing video-level anomaly decisions through structured prompting, the primary limitation is not feasibility, but reliability. In particular, current models exhibit a strong conservative decision tendency in zero-shot settings: they default toward “normal,” rarely flagging abnormal events without explicit guidance. This behavior highlights a fundamental challenge for deployment in open-world surveillance, where anomalies are rare, context-dependent, and often subtle.

Beyond establishing this gap, our study points to two practical axes that meaningfully shape model behavior: temporal context and categorical prompting. Adjusting clip duration can change sensitivity by altering the amount of evidence available for a decision, but longer context does not universally translate into better detection and may plateau depending on scene complexity and data characteristics. More importantly, providing class-specific guidance consistently shifts how the model interprets events, offering a structured semantic frame that reduces ambiguity and encourages anomaly signaling. These observations suggest that progress in MLLM-based VAD will depend less on simply increasing data fidelity and more on improving decision calibration, reasoning robustness, and sensitivity under uncertainty. Future work should therefore prioritize recall-aware prompting strategies, better alignment of anomaly definitions with context, and evaluation protocols that reflect the decision-boundary requirements of real surveillance systems.
\section*{Acknowledgments}

This research is supported by the National Science Foundation (NSF) Award Number 2329816.

\bibliographystyle{ieeetr}  
\bibliography{main} 

\end{document}